# Modeling and forecasting Spread of COVID-19 epidemic in Iran until Sep 22, 2021, based on deep learning

Jafar Abdollahi[a], Amir Jalili Irani[b], Babak Nouri-Moghaddam[a,c,*]

[a] Department of Computer Engineering, Ardabil Branch, Islamic Azad University, Ardabil, Iran
[b] Sama technical and vocational training college, Islamic Azad University, Ardabil branch, Ardabil, Iran
[c] Department of Industrial Engineering, Iran University of Science and Technology, Tehran, 1684613114, Iran

**Abstract**

The recent global outbreak of covid-19 is affecting many countries around the world. Due to the growing number of newly infected individuals and the health-care system bottlenecks, it will be useful to predict the upcoming number of patients. This study aims to efficiently forecast the is used to estimate new cases, number of deaths, and number of recovered patients in Iran for 180 days, using the official dataset of the Iranian Ministry of Health and Medical Education and the impact of control measures on the spread of COVID-19. Four different types of forecasting techniques, time series, and machine learning algorithms, are developed and the best performing method for the given case study is determined. Under the time series, we consider the four algorithms including Prophet, Long short-term memory, Autoregressive, Autoregressive Integrated Moving Average models. On comparing the different techniques, we found that deep learning methods yield better results than time series forecasting algorithms. More specifically, the least value of the error measures is observed in seasonal ANN and LSTM models. Our findings showed that if precautionary measures are taken seriously, the number of new cases and deaths will decrease, and the number of deaths in September 2021 will reach zero.

**Keywords:** LSTM models. Coronavirus, ARIMA, Autoregressive, forecast COVID-19.

## 1. Introduction

At the end of December, the 2019 novel Coronavirus disease (COVID-19) first appeared in Wuhan, Hubei Province, China, and rapidly spread all over the world in a few weeks. On January 30, 2020, with the outbreak of Covid epidemic 19, the Director-General of the World Health Organization, Dr. Tedros Adenom, declared public health as an emergency concern of international concern (Guo YR et al. (2019)), According to the latest WHO and Ministry of Health and Medical Education (MoHME) report on Sep 14, 2021, there were 120,241,488 confirmed cases and 2,662,303 deaths in worldwide and ١,739,360 confirmed cases, and 61,142 deaths in Iran, respectively. Regarding the rapid transmission rate between people, all nations attempt to save their lives by implementing control measures such as travel restrictions, quarantines, social distancing, and hard and soft locks, as well as postponement and cancellation of events (Acter T et al. (2020)). Therefore, it is vital to assess the effectiveness of those control measures on epidemic progression in favor of worldwide expectations, and modeling-based forecasting can assist in health resource management and planning for prevention purposes. (Hu Z et al. (2020)), and (Yang Z et al. (2020)).

To predict additional resources to combat the epidemic, mathematical and statistical modeling tools can be useful for generating timely short-term predictions of reported cases. These forecasts could include estimates of expected rainfall, which could help guide public health officials in providing medical care and other resources needed to fight the epidemic. Short-term predictions can also guide the severity and type of interventions needed to reduce the epidemic. According to promising news about the development of Covid-19 vaccine by the United States and Russia, if this vaccine is not available to Iran by the first decade of 2021, effective implementation of coronary restrictions and personal protection and observance of health protocols and social distance to control the epidemic is vital. (Roosa, K et al. (2020)).

Predicting time series data is an important issue in economics, business, and finance. Traditionally, there have been several methods for effectively predicting the subsequent latency of time series data, such as univariate Autoregressive (AR), univariate Moving Average (MA), Simple Exponential Smoothing (SES), and more notably Autoregressive Integrated Moving Average (ARIMA) with many variations, In particular, the ARIMA model has demonstrated its performance in accurately predicting future time series delays. With recent advances in the computing power of computers and, most importantly, the development of more advanced machine learning algorithms and approaches such as deep learning, new algorithms for analyzing and predicting time series data have been developed. (Siami-Namini, S et al. (2018))

Short-term memory networks or LSTMs for short can be used to predict time series. Different types of LSTM models can be used by changing the parameter according to the data set for each specific type of time series prediction problem (Siami-Namini, S et al. (2018)). In a self-centered model, we predict the desired variable using a linear combination of the variable's past values. The term regression itself indicates that it is a variable return to itself... It is like multiple regressions but with backward values of yt as predictors (Marple, L. (1980)). ARIMA, short for "Autoregressive Integral Moving Average", based on the entered time-series information, can be used alone to predict future values (Nury, A. et al. (2017))

Artificial neural networks are predictive methods based on simple mathematical models of the brain. They enable complex nonlinear relationships between the response variable and its predictors. (Nikam, S. S. (2015))

The novel Coronavirus (COVID-19) has spread significantly around the world and poses new challenges for the research community. Although governments are implementing countless social containment measures, the need for health care systems has increased dramatically and effective management of infected patients is becoming a challenging problem for hospitals. Therefore, accurate short-term prediction of the number of newly infected and recovered cases is crucial to optimizing existing resources and arresting or slowing the progression of such diseases. Recently, deep learning models have shown significant improvements when processing time-series data in various applications. (Zeroual, A et al. (2020))

This paper presents a comparative study of four in-depth learning methods to predict the number of new cases and cases of improvement. In particular, Prophet, Long short-term memory (LSTM), Autoregressive (AUTO REG), Autoregressive Integrated Moving Average (ARIMA) Models. Artificial neural networks (ANNs) and multiple regression algorithms have been used to predict COVID-19 cases globally based on small amounts of data.

This study is based on daily confirmed and improved cases from Iran. The results show the promising potential of the deep learning model in predicting COVID-19 cases and highlight the superior performance of LSTM and ANN compared to other algorithms.

However, in emergencies, even the simplest model can be very complex, and small gaps between different areas of research can be like holes. Therefore, in this paper, we present a simple iterative method for predicting the number of COVID-19 cases, assuming that government data is legitimate and true. The goal is not to strive for precision, but to present our method as the ultimate level of art, but merely to provide the first views and guidelines on the basics. We will be happy if our work anticipates further research to provide accurate and precise methods. The objectives of the article are stated in the following order:

i. Significant improvement in forecast accuracy
ii. Achieve a high level of reliability in the forecast.
iii. Use multiple models to increase the estimate of the final model
iv. Increased accuracy and reduced error compared to single-core models

**So the main contribution of this article:**

i. Developed deep learning methods to forecast the COVID-19 spread.
ii. Four deep learning models have been compared for COVID-19 forecasting.
iii. Time-series COVID19 data from Iran are used.
iv. Results demonstrate the potential of deep learning models to forecast COVID-19 data.
v. Results show the superior performance of the LSTM model.

**The remaining of this article is mainly described as follows**. Section 2 reviews the past work in machine learning for the diagnosis of chronic diseases. The proposed method is introduced in detail in Section 3. Section 4 presents the case study. Finally, the Discussion and conclusion summarize the paper and propose future work in Sections 5 and 6.

## 2. Related Work

Based on real-time data (Dong E, et al. (2020)), confirmed cases of Coronavirus in 2019 (COVID-19) are growing exponentially in most countries of the world. In Italy and Spain, the current epidemic is overburdening the healthcare system (Remuzzi A et al. (2020)), and if the current trend continues, it will not be long before this bitter reality becomes a reality in many European countries and the United States. Become. Therefore, the prediction of COVID-19 release plays a key role (Li Q et al. (2020)), and (Lai A et al. (2020)). In the first place, informing governments and health care professionals of the necessary expectations and actions, and secondly, motivating the community to adhere to the measures that have been put in place to slow down the spread, lest a deplorable scenario occurs. (Ippolito G et al. (2020)), and (McCloskey B et al. (2020))

Research on epidemic processes in statistical physics has a long and fruitful history (Pastor-Satorras R et al. (2015)), and (Wang Z et al. (2016)). Simple mathematical models that describe the essence of epidemic spread can be used to place data with a measurable number of parameters and the values obtained can be used to make informed predictions. In recent years, the research community has also gathered a great deal of evidence in favor of complex and heterogeneous connection patterns in social networks (Boccaletti S et al. (2006)), and (Lü L et al. (2016)). These play a key role in determining the behavior of equilibrium and non-equilibrium systems in general, and the spread of epidemics, and finding optimal control strategies in particular.

Interdisciplinary discoveries at the interface of statistical physics, network science, and epidemiology, documented by large amounts of our health and lifestyle data, have led to digital epidemiology (Salathe M et al. (2012)) and the theory of epidemic processes in complex networks [18]. From classical models that assume completely different populations to newer models that make up the behavioral feedback and structure of our social networks, we have come a long way in better understanding disease transmission and disease dynamics. We can now use this knowledge to develop effective prevention strategies (Wang Z et al. (2016)) and more broadly, we can use the synergies between these different areas of research to improve our lives and communities. (Helbing D et al. (2015)), and (Perc M et al. (2019))

Therefore, in addition to vaccination or diagnostic tests, mathematical modeling can provide an appropriate tool for forecasting and selecting suitable intervention strategies for rapid control of communicable diseases (Vaishya R et al. (2020)), and (Zareie B el al. (2020)). Many different mathematical models are used to predict infectious diseases. The most current mathematical model used for predicting the spread of infectious diseases is the susceptibility-to-epidemic model and its variants, which used several findings from this model for coronavirus prediction ((Yang Z et al. (2020)), Zareie B el al. (2020)), and (Zhan C et al. (2020))). There is no doubt. Findings by this mathematical modeling can be used to manage infectious diseases such as SARS-Cov2.

However, there are many limitations (Wang J. (2020)). The majority of the "Data-Driven" approaches (Knight GM et al. (2016)). such as moving average autoregressive (ARIMA), moving average (MA), and autoregressive (AR) used in previous studies were needed with linear data and such methods are difficult to predict real-time transmission rate of infectious diseases like SARS-Cov2. To model the transmission dynamics of the current COVID-19 outbreak, a broad variety of statistical and mathematical models (Benvenuto D et al. (2020)), and (Dehesh T et al. (2020)) have been used. However, due to the nonlinearity of the new coronavirus data, these models are not accurate enough to predict the rapid transmission rate of the virus.

Recently, as a result of this global crisis, the medical industry was looking for new technologies to combat the COVID-19 outbreak. Artificial Intelligence (AI) is one of the most successful factors for technology in healthcare systems. The AI easily tracks its prevalence, identifies at-risk patients, and helps with real-time monitoring of the infection. It can also predict mortality through an adequate analysis of previous data. It can also help eliminate the spread of the virus through population screening, health care, information, and infection control instructions (Hu Z et al.(2020)).

Therefore, studies have focused on effective methods to meet the challenges of the mathematical models for the worldwide COVID-19 epidemic prediction, such as Deep Learning Network (Yudistira N. (2020)), and (Huang C-J et al. (2020)).

2.1 Forecast Model

Prediction analysis tools are provided by several different models and algorithms that can be applied in a wide range of applications. Determining which best predictive modeling practices are best for your company is making the most of a predictive analytics solution and using data for key insightful decisions. One of the most widely used models of forecast analysis is the forecast model in metric value forecasting, estimating the numerical value for new data based on what has been learned from historical data. This model can be applied to any place where historical numerical data are available. (Sarkar, K. et al. (2020)), (Al-Qaness et al. (2020)), and (Fanelli, D et al. (2020)).

## 3. Material and Methods

### 3.1. Data source

We have daily obtained updates from the website of the National Health Commission of Iran (national Health Commission of Iran) on the cumulative number of confirmed cases reported for the 2019-nCoV epidemic throughout the provinces of Iran. Data updates are collected daily at noon (GMT-5), between 22 February 2020 and 12 March 2021. The short-run series is affected by reporting irregularities and delays, so cumulative curves are more stable and are likely to provide more stable and reliable estimates. Therefore, we analyze the cumulative path of the epidemic in Iran.

The provided dataset by MoHME( Https://behdasht.gov.ir/) is used to feed our model. Shown in Tables 1 and 2. This dataset contains confirmed statistics about patients with COVID- in the provinces of Iran from February 2020 to March 2021, including three sections as several deaths, recovered patients, and has been confirmed to date.

**Table 1**
Features of Covid-19 type dataset

| Input | Number |
| --- | --- |
| Date | From 25 Feb 2020 to 12 March 2021 November ( 12 Mount) |
| Confirmed | 381 Record |
| Deaths | 381 Record |
| Recovered | 381 Record |

This data set contains 381 records and four features such as date, Confirmed, Death, and Recovered. This data set has been collected and processed from February 25, 2020, to March 12, 2021 (within one year) according to the official daily statistics through the Ministry of Health, Treatment, and Medical Education of Iran.

**Table 2**
Description of the Covid-19 datasets.

| Description of the Covid-19 Datasets. | | | | |
| --- | --- | --- | --- | --- |
| Dataset | Sample Size | Feature size including class label | Presence of missing attribute | Presence of noisy attributes |
| Datasets | 381 | 4 | NO | NO |

### 3.2. Data Analysis

The flow chart of the methodology applied in this study is illustrated in fig. 1. A detailed description of each step will be discussed in the following.

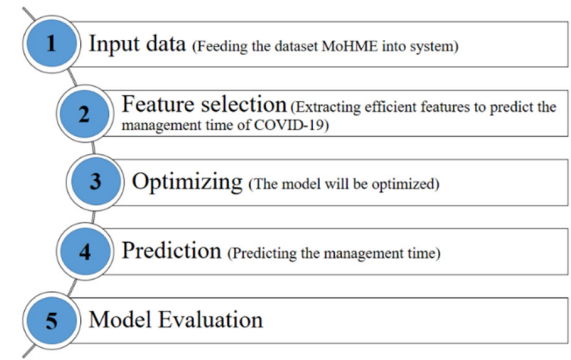

**Fig. 1.** The flow chart of the implemented method

### 3.3 Model

Algorithms: an overview

Fig. 2. shows a flow chart of the algorithms we applied in this study. It shows how our hybrid system is designed. We generate short-term forecasts in real-time using four phenomenological models that have been previously used to derive short-term forecasts for several epidemics for several infectious diseases, including SARS, Ebola, pandemic influenza, and dengue.

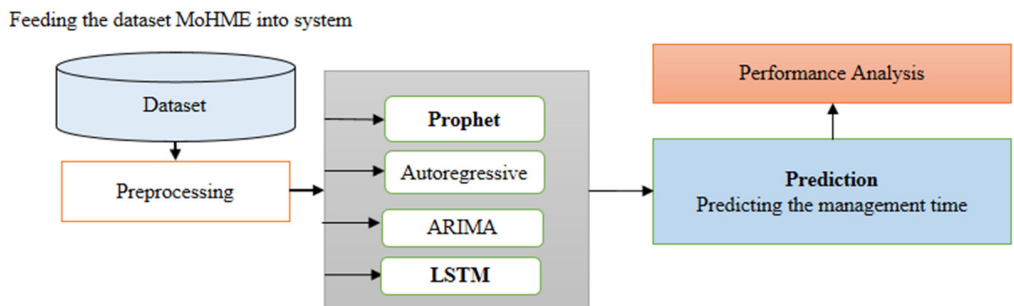

**Fig. 2.** A schematic flow chart of the algorithms.
Different methods have been utilized mostly based on machine learning techniques for forecasting covid-19. In this paper, a new method has been performed on the Covid-19 database to investigate the ability of time series techniques for Forecasting covid-19 In this paper, a new method has been performed on the

Covid-19 database to investigate the ability of time series techniques for Forecasting covid-19. To this end, the data related to covid-19 have been preprocessed. In the sequel, four-time series analysis and machine learning techniques were developed based on including Voting to make forecasting over the database. The machines have been evaluated using the MSE technique.

3.4. Evaluation

The most common evaluation criteria for predicting RMSE. MAPE because it is scale-independent and shows the ratio of error to actual values as a percentage. And MASE, which shows how well the forecast performance performs compared to the average simple forecast. The mean squared error (MSE) is largely used as a metric to determine the performance of an algorithm. The formula to calculate the MSE is as follows:

$$MSE = \frac{\sum_{i=1}^{n}(y_i - y_i)^2}{n} \tag{1}$$

$$MSE = \frac{1}{n}\sum_{i=1}^{n}(y_i - y_i)^2 \tag{2}$$

Defining the variables
1. n - the total number of terms for which the error is to be calculated
2. $y_i$ - the observed value of the variable
3. $y_i$ the predicted value of the variable

The mean square error is the average of the square of the difference between the observed and predicted values of a variable.

4. Experiment

Artificial intelligence (AI) is being developed and deployed around the world in a variety of fields. The World Health Organization (WHO) declared the coronavirus and epidemic on March 14, 2021 citing more than ١٢٠ Million cases in more than ٢٠٢ countries and territories around the world at the time. At the time of writing, the number of confirmed cases has quickly exceeded 120,104,930 underscoring the sustained risk of further global expansion. Governments around the world are implementing various containment measures as the health care system prepares for a tsunami of infected people seeking treatment. Therefore, it is important to know what to expect in terms of growth in the number of cases and to understand what is needed to capture very worrying trends.

For this purpose, we show here the predictions obtained by four simple iteration methods that require only the daily values of the verified items as input. This method takes into account the expected improvements and deaths and determines the maximum allowable daily growth rate, which leads to moving away from the exponential increase towards stable and decreasing numbers.

4.1. Outlook

As we hope, the forecasts clearly show epidemic growth is a completely non-linear trend, where every day the loss of inactivity is one day too much. Even with a few days left on the road that we do not operate today, the difference between a manageable situation and a health care system can be overwhelming. The outlook depends a lot on whether we take these facts into account and act on them. Governments can change travel bans, close shops, and restaurants, and encourage us to stay home. Ultimately, however, each of us has a responsibility to respect these limitations and do our best to minimize the possibility of future infections.

Keeping the daily growth rate below at least 5% is an important goal for a promising outlook. Data from China, where the COVID-19 appears to be coming to an end, confirm this advertisement. Around mid-February, the daily growth rate there dropped to about 4% and then to 3% and below. This is the beginning of a confirmed case plateau that, along with recovery and death, leads to a reduction in the number of infected people. Singapore, South Korea, and Hong Kong have also successfully changed their epidemic by using the strict tactics used in China. Unfortunately, this has not been the case in many countries. (Cohen J et al. (2020))

We have two options. The first is to limit our behavior so that Covid-19 is as fast as it grows. The second is that we allow it to slip so that the situation becomes so dire that the rulings of the government force us to limit our behavior. There is still time to act, but a rosy landscape is rapidly disappearing. In recent global emergencies, scientists, physicians, and healthcare professionals around the world are continuing to search for a new technology to support the Covid-19 pandemic. (Lalmuanawma, S et al. (2020))

Evidence from the Machine Learning (ML) and Artificial Intelligence (AI) program of previous epidemics encourages researchers to provide a new perspective on combating coronavirus outbreaks. Artificial intelligence (AI) is an innovative technology that is useful in combating the COVID-19 epidemic. Covid-19 artificial intelligence technology is useful for properly screening, tracking, and predicting current and future patients and diagnosing infection. (Abdollahi, J. et al. (2019)), and (Abdollahi, J. et al. (2020)).

## 5. Results

In a study by Zareie B el al. (2020)), the Generalized Additive Model was used to predict the coronavirus epidemic in Iran based on China Parameters. One of the main drawbacks of this analysis was the ambiguity of the data from the first day of the Iran epidemic so that they could not identify a better model with AIC or BIC criteria to choose the best predictor subsets in regression. Before implementing the data, I will divide it into a set of training and forecasting/testing. So far, I have made my choice in the ARIMA model based only on the training set. One of the most widely used methods in time series prediction is ARIMA (Autoregressive Integrated Moving Average).

However, the Arima algorithm has a weakness in finding the optimal model. Forecast Validation operator with the Arima operator in the training and a Performance (regression) operator in the testing subprocess inside and outside and Optimize operator. First, connected the performance output port of the Forecast Validation with the performance port of the Optimized. Then selecting p, d, q for optimization. I determine the ARIMA parameters p and q to run from zero too. This choice of range is arbitrary. Here the de parameter is allowed to be zero or one. Through this, we test the performance of time series methods and store the results. Whereas the predicted results with ARIMA optimization resulted in MSE 0.029 From these results, it can be concluded that the optimization of ARIMA produces a better forecasting model than the ARIMA model without optimization.

Also, we created two LSTM layers using the BasicLSTMCell method. Each of these layers has several units defined by the parameter num_units. Apart from that, in this work, we used to combine layers in a network. Then we used the static_rnn method to construct the network and generate the predictions. We saw two approaches when creating LSTM networks. Both methods had simple problems, and each used a different API. This way one could see that TensorFlow is more detailed and flexible, however, you need to take care of a lot more stuff than when you are using Keras. Kera's simpler than TensorFlow and does not have the features that TensorFlow provides, but both methods work well and can work better.

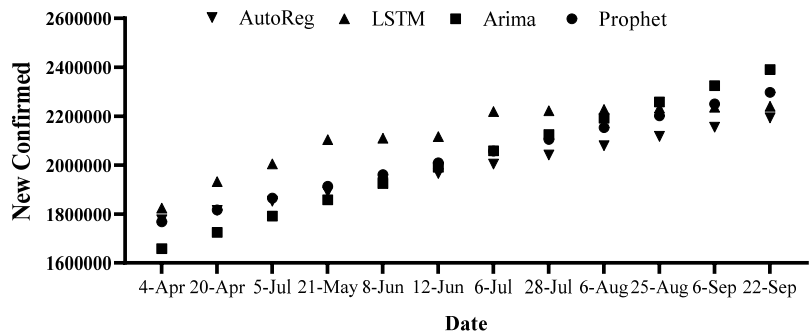

**Fig. 3**. Number of Confirmed in the next quarter

In Fig 3 - **_Prophet,_** Long short-term memory (**_LSTM_**), Autoregressive (AUTO REG), Autoregressive Integrated Moving Average (ARIMA).

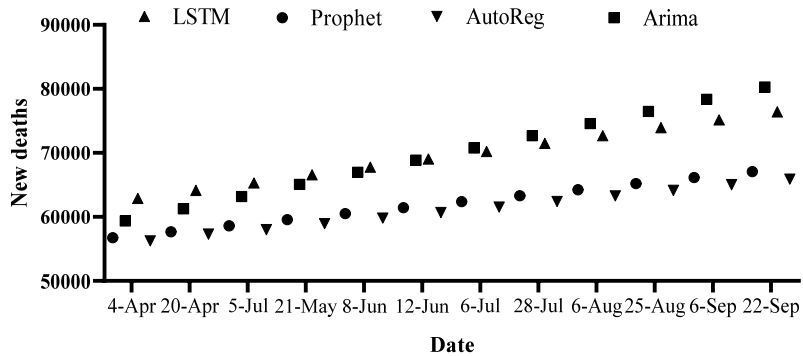

**Fig. 4.** Number of deaths in the next quarter

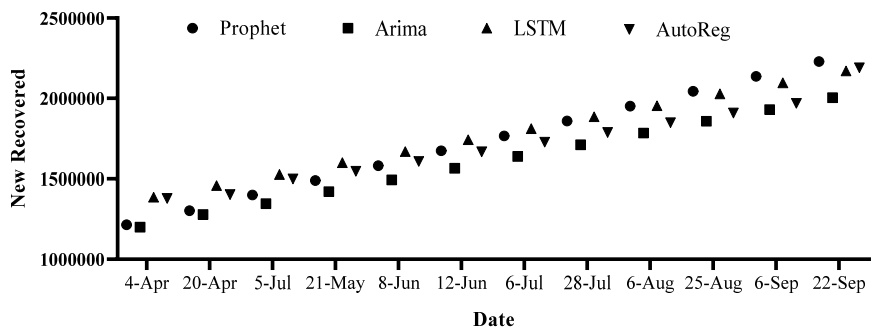

**Fig. 5.** Number of Recovered in the next quarter

**Table 3**
describe the MSE value of each method

|  | Prophet | LSTM | AUTO REG | ARIMA |
|---|---|---|---|---|
| Train Score | 0.76 | 0.86 | 0.82 | 0.75 |
| Test Score | 0.81 | 0.85 | 0.81 | 0.69 |
| MSE Train | 0.051 | 0.018 | 0.037 | 0.056 |
| MSE Test | 0.052 | 0.011 | 0.091 | 0.029 |

Given the outbreak of the coronavirus, health policymakers in Iran have implemented various plans such as closing the specific centers, locations, institutions, and physical distancing. Although, in the early days of the epidemic, Iranian people followed many of the above-mentioned plans. Recently, however, many people didn't follow control measures due to the economic problems they have. On the other hand, it has become normal for the Iranian people.

This study aimed to achieve a model to predict Coronavirus statistics in terms of total confirmed cases, total deaths, and recovered patients in two cases, including with and without control measures. Using a mask, avoiding touching the face, maintaining social distance, and washing hands regularly is examples of control measures. The results illustrated in Fig. 3-5 implies that if these controls are not taken seriously, an increase in the number of new cases and the mortality rate is predictable.

## 6. Discussion

The current models show that the upcoming next few months will be hard for the world. The control system adopted by the various national governments is in fact very strict and works well. In addition, adopting a direct posture can effectively monitor improved patients and control patient mortality. The findings of this study may explode if the government does not take strict control measures for its residents. Arranging emergency clinics and improving clinical offices should be done as soon as possible to prevent the development of the country to prevent from happening. We have used several methods to observe mortality from COVID-19. Data is unstable.

By looking at figure number (3–5), it is difficult to prove which method is suitable for this time series dataset in future predictions. To overcome this situation, we describe the MSE value of each method in Table 3. Compared with other methods, the LSTM method has a lower MSE score. Therefore, the LSTM method is suited to all other methods. In the LSTM model, using grid search, we identified a set of parameters that produced the best-fit model for our time series data. By continuing the model, future predictions of death cases indicate that the number of deaths will increase by 65,905 to more than 76,403 by September 22, 2021 and beyond. But governance parameters, weather conditions, state-level population, and national geographical distribution may affect the forecast. This can further improve the predictability of the model.

## 7. Conclusion and future work

The empirical studies conducted and reported in this article show that deep learning-based algorithms such as ANN and LSTM outperform traditional-based algorithms such as the ARIMA model. More specifically, the average reduction in error rates obtained by LSTM was between 84 — 87 percent when compared to time series algorithms indicating the superiority of RNN to other time-series Algorithms. During the implementation process, it was observed that the number of training, under the heading of "epoch" in deep learning techniques, affects the performance of the model. In conclusion, the outbreak can be controlled by attention to the control measures, which represents the positive effect of observing protocols by highlighting the fact that both mortality and the new infected rate are decreasing day by day.

**Further, works**: We are trying to use artificial intelligence to analyze images taken from patients to identify Tuberculosis. It can significantly help to automate the scanning process and also change the form of workflow with minimal contact with patients and the protection of medical imaging technicians. It can also improve work efficiency and facilitate subsequent quantities by identifying infection accurately in X-ray and CT images.

**Conflict of interest**
The authors declare that no conflict of interest exists.
**Acknowledgment**
We are thankful to our colleagues who provided expertise that greatly assisted the research.
**The source of funding**
All the funding of this study was provided by the authors.